\documentclass[a4]{article}

\usepackage{arxiv}

\usepackage[utf8]{inputenc} 
\usepackage[T1]{fontenc}    
\usepackage{hyperref}       
\usepackage{url}            
\usepackage{booktabs}       
\usepackage{amsfonts}       
\usepackage{nicefrac}       
\usepackage{microtype}      
\usepackage{natbib}

\usepackage[utf8]{inputenc}
\usepackage{amsmath}
\usepackage{bbm}
\usepackage{graphicx}
\usepackage{amssymb}
\usepackage{url}
\usepackage{mathtools}
\usepackage{flexisym}
\usepackage[inline]{enumitem}
\usepackage{hyperref}
\usepackage{booktabs}
\usepackage[flushleft]{threeparttable}
\usepackage[super]{nth}
\hypersetup{
    colorlinks=true,
    urlcolor=blue
    }
\newcommand{\etal}{\textit{et al.}}
\usepackage[dvipsnames]{xcolor}
\usepackage[labelfont=bf]{caption}
\usepackage{multirow}
\usepackage{mfirstuc}
\MFUnocap{in}
\MFUnocap{the}
\MFUnocap{as}
\MFUnocap{it}

\title{\normalfont Does CLIP Benefit Visual Question Answering in the\\ Medical Domain as Much as it Does in the General Domain?}

\author{Sedigheh Eslami \\
	D4L data4life gGmbH\\
	Potsdam, Germany \\
	\And
	Gerard de Melo\\
	Hasso Plattner Institute\\
	Potsdam, Germany \\
	\And
	Christoph Meinel\\
	Hasso Plattner Institute\\
	Potsdam, Germany \\
	\And
	\texttt{sedigheh.eslami@data4life.care} \\
	\texttt{\{gerard.demelo, christoph.meinel\}@hpi.de}
}

\renewcommand{\headeright}{}
\renewcommand{\undertitle}{}
\renewcommand{\shorttitle}{Does CLIP benefit visual question answering in the medical domain as much as it does in the general domain?}

\begin{document}
\maketitle

\begin{abstract}
\textbf{Objective:} Contrastive Language--Image Pre-training (CLIP) has shown remarkable success in learning with cross-modal supervision from extensive amounts of image--text pairs collected online. Thus far, the effectiveness of CLIP has been investigated primarily in general-domain multimodal problems. This work evaluates the effectiveness of CLIP for the task of Medical Visual Question Answering (MedVQA). To this end, we present PubMedCLIP, a fine-tuned version of CLIP for the medical domain based on PubMed articles.\\
\textbf{Materials and Methods: } Our experiments are conducted on two MedVQA benchmark datasets and investigate two MedVQA methods, MEVF (Mixture of Enhanced Visual Features) and QCR (Question answering via Conditional Reasoning). For each of these, we assess the merits of visual representation learning using PubMedCLIP, the original CLIP, and state-of-the-art MAML (Model-Agnostic Meta-Learning) networks pre-trained only on visual data. We open source the code for our MedVQA pipeline and pre-training PubMedCLIP.\\
\textbf{Results:} CLIP and PubMedCLIP achieve improvements in comparison to MAML's visual encoder. PubMedCLIP achieves the best results with gains in the overall accuracy of up to $3\%$. Individual examples illustrate the strengths of PubMedCLIP in comparison to the previously widely used MAML networks.\\
\textbf{Discussion and conclusion:} Visual representation learning with language supervision in PubMedCLIP leads to noticeable improvements for MedVQA. Our experiments reveal distributional differences in the two MedVQA benchmark datasets that have not been imparted in previous work and cause different back-end visual encoders in PubMedCLIP to exhibit different behavior on these datasets. Moreover, we witness fundamental performance differences of VQA in general versus medical domains.
\end{abstract}


\section{BACKGROUND AND SIGNIFICANCE}
Medical visual question answering (MedVQA) is the task of answering natural language questions about a given medical image. To solve such multimodal tasks,
a system must interpret both visual and textual data as well as infer the associations between a given image and a pertinent question sufficiently well to elicit an answer \cite{antol2015vqa}. The development of MedVQA has considerable potential to benefit healthcare systems, as it may aid clinicians in interpreting medical images, obtaining more accurate diagnoses by consulting a second opinion, and ultimately, may expedite and improve patient care. Achieving this in the medical domain in particular is non-trivial, as we suffer from a general lack of sufficient and balanced training data. The ImageCLEF community hosts annual MedVQA challenges \cite{vqa-med2020, vqa-med2021}, where new VQA datasets using PubMed articles are released. However, there are concerns about whether the question--answer pairs in these datasets are realistic and clinically relevant \cite{rad}. For example, in the VQA-Med 2021 Challenge \cite{vqa-med2021}, the dataset consisted entirely of questions asking about the category of abnormality in the image. This lack of diversity in the semantics of the questions meant that the winning teams were able to treat the MedVQA problem as a multi-class image classification task, without any need to interpret the questions \cite{gong2021, eslami}. Lau \etal \cite{rad} published VQA-RAD as the first public benchmark dataset comprising realistic and clinically relevant question--answer pairs generated by expert clinicians and radiologists. Recently, Liu \etal \cite{slake} created the bilingual SLAKE dataset that includes not only clinically relevant data, but also mask and bounding box annotations for images, which are beneficial for semantic segmentation and detection of organs in medical images.

Current approaches for this multimodal task adopt deep neural encoders to interpret the image and the question and then pick a corresponding answer. They typically consist of four main components: a visual encoder, question encoder, attention-based fusion of vision and text features, and an answer classifier \cite{qcentric, QCR, MEVF, muvam, miccai2021liu}. Skip-thought vectors, LSTM, and GRU recurrent neural networks have been popular question encoders in prior work. Bilinear attention networks \cite{ban}, stacked attention networks \cite{san}, and element-wise production are popular as multimodal pooling approaches in MedVQA. With regard to the visual encoder, the majority of previous MedVQA papers \cite{muvam, QCR, MEVF} employ the Mixture of Enhanced Visual Features (MEVF) \cite{MEVF}. It consists of two modules: 
\begin{enumerate*}
    \item the pre-trained meta learning module, which uses Model-Agnostic Meta-Learning (MAML) \cite{MAML} with the objective of solving a $k$-shot $n$-way classification problem with the abnormality status of different organs as classes,
    \item the Convolutional Denoising Autoencoder (CDAE) \cite{CDAE} module in order to denoise the medical image.
\end{enumerate*}
However, the pre-training of MEVF is custom-tailored for the particular challenges encountered in the VQA-RAD benchmark dataset and is specifically designed for the organs present in this dataset, i.e., the chest, brain, and abdomen, limiting its generalizability to other settings. Liu \etal \cite{miccai2021liu} similarly restricted the objective of their visual encoding to chest, brain, and abdomen, and pre-train three separate visual feature extraction teacher models for these respective body regions. Furthermore, they distilled the three teacher models into a smaller student model by contrastive representation distillation. This motivated us to design an alternative model, PubMedCLIP, which learns features in medical images of various body organs and is not limited to only a few regions.

Transfer learning and making use of pre-trained models has become an inseparable part of representation learning in computer vision and natural language processing. Recent work \cite{CLIP, VLBART, vlbert} has shown improvements of visual and textual encoders when learning from the contrast of image--text pairs and using natural language as supervision in addition to just visual images. This trend of improvements has also been observed in various classification use cases in the medical domain \cite{contrastive-manning}. 
Among these approaches, the contrastive pre-training of language--image data in CLIP \cite{CLIP} has been particularly successful. CLIP was trained using a very large number of image--text pairs acquired from the Internet with close to zero additional human annotation. This aspect is particularly useful for the medical domain, since data annotation requires expert medical knowledge and thus is often an expensive and time-consuming obstacle. Following CLIP, we investigate to what extent using publicly available medical image--text pairs without any further annotation can be useful for the MedVQA task. To this end, we consider a large number of medical image--text pairs obtained from PubMed articles and use them to train PubMedCLIP. We further examine the outcomes when incorporating PubMedCLIP as the visual encoder into state-of-the-art MedVQA methods. We investigate whether fine-tuning CLIP for the MedVQA task benefits medical VQA as much as it benefits general-domain VQA, as observed in previous work \cite{CLIPbenefit}. 

To the best of our knowledge, this is the first study introducing a PubMed-optimized CLIP and assessing the effects of using CLIP for MedVQA. In contrast to previous visual encoders used in MedVQA, PubMedCLIP is pre-trained using medical images from a diverse range of body regions and is not restricted to only brain, chest, and abdomen images. We conduct extensive experiments on two MedVQA benchmark datasets and employ diverse back-end visual encoders in PubMedCLIP. Our experiments reveal that using PubMedCLIP as a pre-trained visual encoder improves previous models by up to $3\%$.

\section{MATERIALS AND METHODS}
\subsection{PubMedCLIP}
Our first step was to consider the original CLIP, which has been pre-trained on general-domain images encountered online, and fine-tune it using medical image--text pairs. To this end, we drew on the Radiology Objects in COntext (ROCO) dataset \cite{ROCO}, which provides over $80K$ samples. ROCO includes diverse imaging modalities such as ultrasound, X-Ray, fluoroscopy, PET scans, mammography, MRI, angiography, from various human body regions, e.g., head, neck, jaw and teeth, spine, chest, abdomen, hand, foot, knee, and pelvis. The image--text pairs in this dataset are captured from PubMed articles. The texts here are taken from the relatively short captions (average length of 20 words) associated with images in the articles, which provide rich explanatory information about the content of images. To the best of our knowledge, ROCO is the only large-scale publicly available medical dataset that includes image--text pairs for a diverse range of body organs and imaging modalities.

In this work, the training and validation data splits from the original paper \cite{ROCO} were used to fine-tune CLIP for the medical domain, with ViT32 Vision Transformer \cite{Dosovitskiy2021ViT}, ResNet RN-50 and RN-50x4 \cite{HeEtAl2016ResNets} visual encoder back-ends. With respect to the maximum text length accepted by CLIP, which is 76, we trimmed any longer captions, while zero-padding shorter ones. 

\begin{figure}
\centering
  \includegraphics[width=\linewidth]{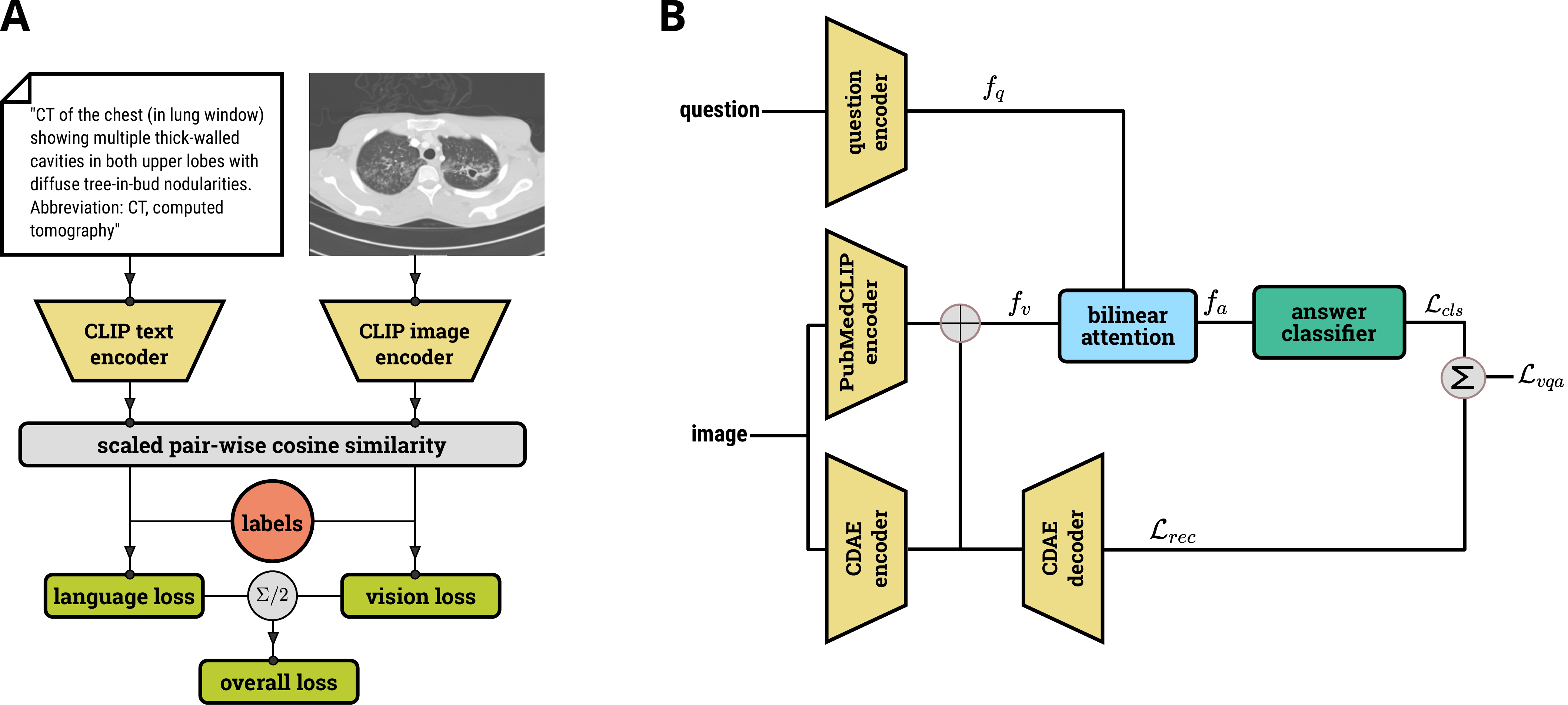}
  \caption{(A) Overview of how PubMedCLIP is pre-trained. (B) Schematic of MedVQA backbone with PubMedCLIP pre-trained visual encoder.}
  \label{fig-pubmedclip}
\end{figure}

We refer to the resulting fine-tuned model as PubMedCLIP. PubMedCLIP was trained for $50$ epochs with a batch size of $64$, and Adam optimization \cite{adam} with a learning rate of $10^{-5}$. The source code along with further implementation details can be found at:
\href{https://github.com/sarahESL/PubMedCLIP/tree/main/PubMedCLIP}{https://github.com/sarahESL/PubMedCLIP}. 
An overview of the training procedure for PubMedCLIP is given in Figure \ref{fig-pubmedclip} (A). Text and image are encoded separately using CLIP. The cosine similarity between text and image features is computed. Finally, the vision and language cross-entropy loss values are computed and their average is considered as the overall loss value.

\subsection{PubMedCLIP in MedVQA}
Our goal is to investigate the effect of using PubMedCLIP as a pre-trained visual encoder in MedVQA models. VQA in this work is considered as a classification problem, where the objective is to find a mapping function $f$ that maps an image--question pair $(v_i, q_i)$ to the natural language answer $a_i$. For our investigation, we considered two prominent MedVQA methods, MEVF \cite{QCR} and QCR \cite{MEVF}, that adopt MEVF as their visual encoders. To assess the contribution of PubMedCLIP, we modified the MEVF by substituting its pre-trained MAML module with PubMedCLIP. Hence, the representative visual feature in our work is the concatenation of the output of the PubMedCLIP network and the CDAE encoder network. Both models use GloVe word embeddings \cite{glove} followed by an LSTM in order to encode questions. Furthermore, the multimodal pooling mechanism for combining question and image features is BAN \cite{ban} in both models. We retained the same question encodings, multimodal fusion, and objective functions proposed for MEVF \cite{MEVF} and QCR \cite{QCR}, respectively, and only replaced the visual encoders. 

The overall objective is to minimize the error of answer classification and image reconstruction, denoted as
\begin{equation}
    \mathcal{L}_{\mathrm{vqa}} =  \mathcal{L}_{\mathrm{cls}} + \mathcal{L}_{\mathrm{rec}}.
\end{equation}
Following previous work \cite{BCELogits}, a sigmoid layer preceding a binary cross-entropy loss computation is used for the classification. The loss function for the autoencoder reconstruction is mean squared error. A schematic architecture of the backbone of our work is shown in Figure \ref{fig-pubmedclip} (B). The answer classifier is a two-layer feed-forward network with the ReLU activation function, as proposed for BAN \cite{ban}.
\subsection{Datasets}
We conducted our experiments using two well-known datasets:

\begin{enumerate}
    \item \textbf{VQA-RAD} \cite{rad} consists of 315 medical images and 3,515 question--answer pairs. We follow previous work by adopting the data split proposed for MEVF \cite{MEVF}. We notice that all the images in the test dataset are also present in the training set. However, the set of question--answer pairs for these images in the test set are unseen in the training set. 
    \item The \textbf{SLAKE} \cite{slake} dataset consists of English and Chinese questions. In this work, we utilize the English subset of the dataset, comprising 642 images and more than 7,000 question--answer pairs. Using the original data split, we observe that in contrast to VQA-RAD, all the images in the test set of SLAKE are unseen in the training set.
\end{enumerate}

\subsection{Experimental setup}
In order to ensure a fair comparison, our experiments generally followed the same setups used in the MEVF and QCR studies. For both methods, Adam optimization was invoked for training. MEVF was trained for 20 epochs, QCR for 200 epochs. When using PubMedCLIP as the pre-trained visual encoder, we set the learning rate to $1\times 10^{-3}$ and $2\times 10^{-3}$ and the batch size to $16$ and $32$ in QCR and MEVF, respectively. All implementations are based on the PyTorch framework \cite{pytorch}. We ran the original MEVF and QCR on our machine and report the results here to have a fair comparison. Due to the non-deterministic behaviour of the cuDNN library used in CUDA convolution operations \cite{DL-nondeterminism}, we observed non-deterministic results in different runs. For a more robust comparison, we repeated all experiments 10 times and report the average accuracy scores.
\section{RESULTS}
\label{sec:exp}

The results of our experiments are reported in Table \ref{table-result}. We provide the overall accuracy along with the accuracy of answering open-end and closed-end questions. It is observed that the performance of both MEVF and QCR approaches are improved when adopting CLIP and PubMedCLIP as the pre-trained visual encoder. Furthermore, results of PubMedCLIP show up to $1\%$ improvement in comparison to the original CLIP. For the VQA-RAD dataset, PubMedCLIP with the ResNet-50 backend achieves the best results, improving the overall accuracy of MEVF up to $6\%$ and for QCR up to $3\%$ percent. Results on the SLAKE dataset indicate that PubMedCLIP with the back-end ViT32 Vision Transformer visual encoder attains the best accuracy. It enhances MEVF up to $3\%$ and QCR up to $2\%$. We witness the same trend of improvement among overall, open-end, and closed-end accuracy scores.

\begin{table*}[htb]
     \centering
     \huge
     \resizebox{\linewidth}{!}{
    \begin{tabular}{llccc|ccc}
     \toprule
     \textbf{MedVQA} &  \textbf{Visual} & \multicolumn{3}{c}{\textbf{VQA-RAD Accuracy}} & \multicolumn{3}{c}{\textbf{SLAKE Accuracy}}
      \\
      \textbf{model} &  \textbf{encoder} & {open} & {closed} & {overall} & {open} & {closed} & {overall} \\
      
    \midrule
    {}  & {MAML + AE}  & {42.1\%} & {73.2\%} & {60.8\%} & {74.1\%} & {77.5\%} & {75.5\%}\\
    \cmidrule(l){2-8}

     \multirow{6}{*}[0em]{MEVF}  & {CLIP-ViT-B + AE}  & {50.8\%} & {75\%} & {65.4\%} & {75.8\%} & {80.5\%} & {77.7\%}\\
    {}  & {CLIP-RN50 + AE}  & {47\%} & {77.4\%} & {65.4\%} & {75.7\%} & {79.6\%} & {77.2\%} \\
    {}  & {CLIP-RN50x4 + AE}  & {46.8\%} & {76.6\%} & {64.8\%} & {75.9\%} & {79.1\%} & {77.2\%} \\
    \cmidrule(l){2-8}
    {}  & {PubMedCLIP-ViT-B + AE}  & {48.9\%} & {76.7\%} & {65.5\%} & \textbf{76.5\%} & \textbf{80.4\%} & \textbf{78\%}\\
    {}  & {PubMedCLIP-RN50 + AE}  & \textbf{48.6\%} & \textbf{78.1\%} & \textbf{66.5\%} & {76.2\%} & {79.9\%} & {77.6\%} \\
    {}  & {PubMedCLIP-RN50x4 + AE}  & {47.1\%} & {77.8\%} & {65.6\%} & {76.6\%} & {79.1\%} & {77.6\%} \\
 \midrule \midrule
    {}  & {MAML + AE}  & {56\%} & {77.9\%} & {69.2\%} & {76.8\%} & {80.6\%} & {78.3\%}  \\
    \cmidrule(l){2-8}
    
    \multirow{6}{*}[0em]{QCR}  & {CLIP-ViT-B + AE}  & {57.6\%} & {79.5\%} & {70.7\%} & {78.6\%} & {81\%} & {79.5\%} \\
    {}  & {CLIP-RN50 + AE}  & {58.3\%} & {80\%} & {71.3\%} & {78.2\%} & {81.5\%} & {79.7\%} \\
    {}  & {CLIP-RN50x4 + AE}  & {59.9\%} & {79.4\%} & {71.3\%} & {77.6\%} & {80.5} & {78.7} \\
    \cmidrule(l){2-8}
    {}  & {PubMedCLIP-ViT-B + AE}  & {58.4\%} & {79.5\%} & {71.1\%} & \textbf{78.4\%} & \textbf{82.5\%} & \textbf{80.1\%} \\
    {}  & {PubMedCLIP-RN50 + AE}  & \textbf{60.1\%} & \textbf{80\%} & \textbf{72.1\%} & {77.8\%} & {81.4\%} & {79.3\%} \\
    {}  & {PubMedCLIP-RN50x4 + AE}  & {60\%} & {79.7\%} & {71.8\%} & {77.7\%} & {81.3\%} & {79.1\%} \\
      \bottomrule
      \end{tabular}}
   \caption{Accuracy scores on VQA-RAD and SLAKE datasets.} \label{table-result}
\end{table*}

\section{DISCUSSION}

\begin{figure}
\centering
  \includegraphics[width=0.7\linewidth]{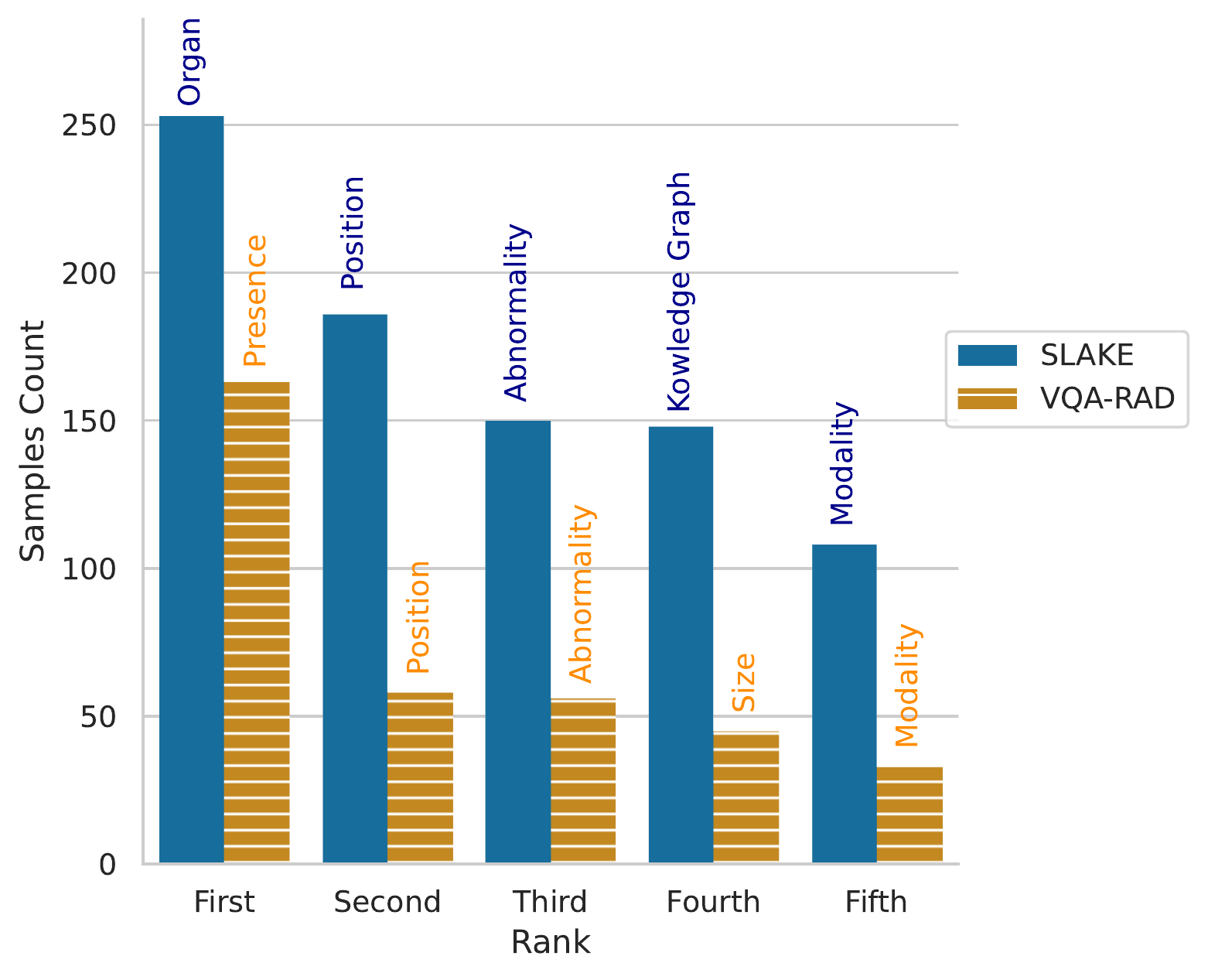}
  \caption{Distribution of the top 5 frequent question types in VQA-RAD and SLAKE datasets.}
  \label{res-plot}
\end{figure}

The fact that ResNet-50 for VQA-RAD and ViT for SLAKE dataset achieve the best results suggests that there are underlying differences in the question type distribution in these datasets. As shown in Figure \ref{res-plot}, the majority of the questions in the VQA-RAD dataset ask about the presence of an abnormality in the images. This requires the visual encoder to detect local features and local abnormalities in the image. Therefore, the CNN-based ResNet model with better visual localization outperforms the Vision Transformer. However, on the SLAKE dataset, the majority of the questions are from the type ``organ", asking which organ is present in the image. For such cases, the visual encoder needs to be able to acquire a holistic overall understanding of the content of the image and thus also capture long-range dependencies of image patches. Vision Transformers indeed are capable of accounting for such features \cite{glance}, and hence perform better on the SLAKE dataset. Figure \ref{res-plot} plots the distribution of question types for the top 5 frequent types in both datasets.


\begin{figure}
\centering
  \includegraphics[width=\linewidth]{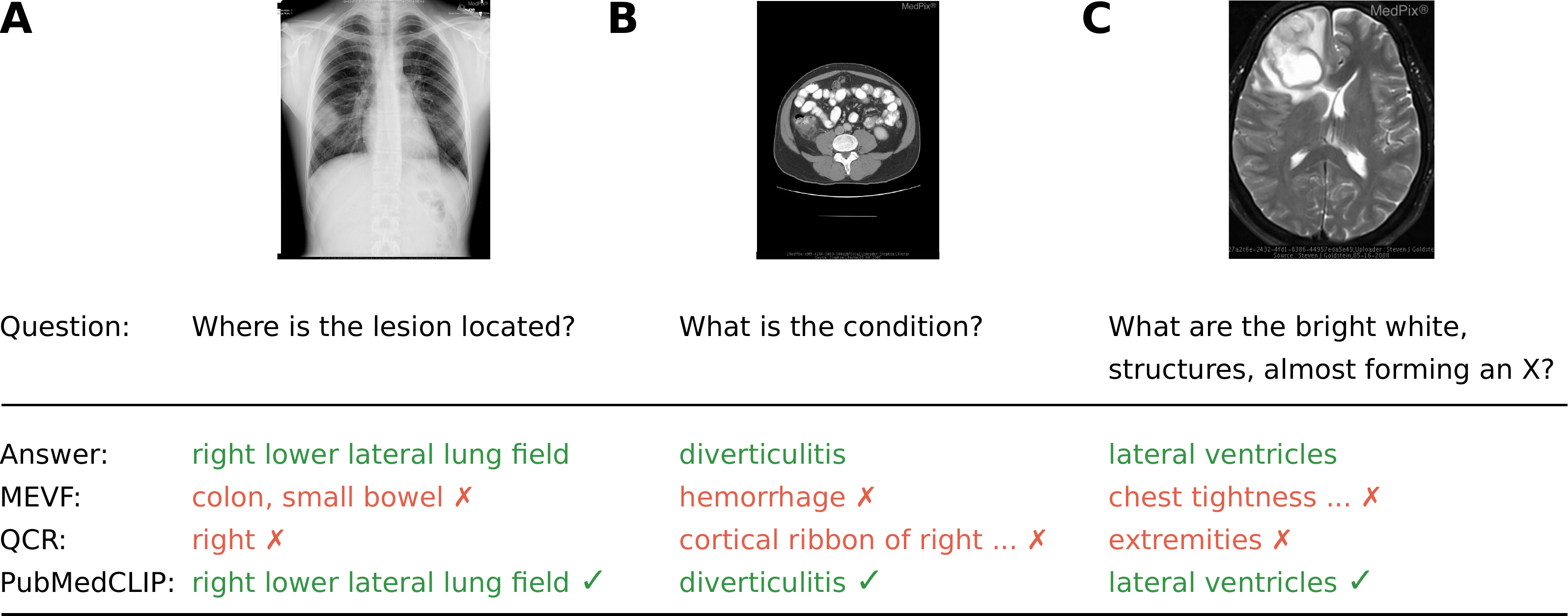}
  \caption{Examples from VQA-RAD dataset.}
  \label{rad-table-example}
\end{figure}

\begin{figure}
\centering
  \includegraphics[width=\linewidth]{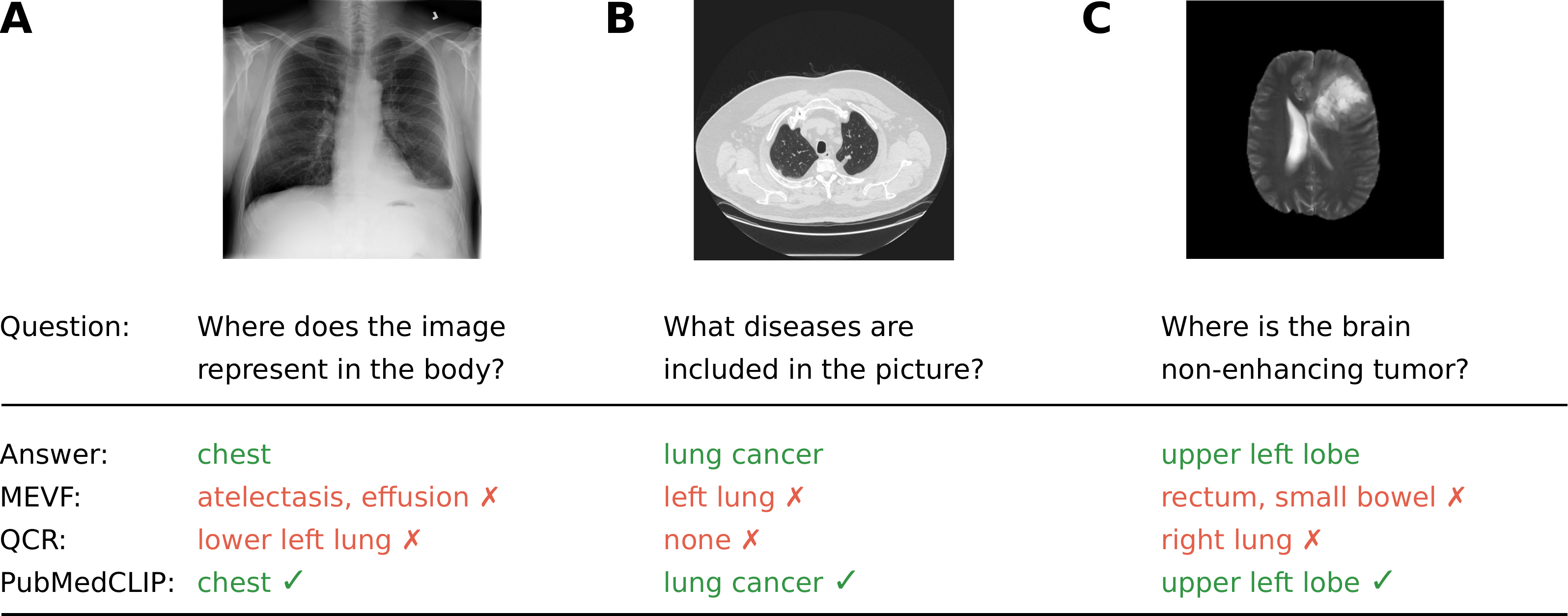}
  \caption{Examples from SLAKE dataset.}
  \label{slake-table-example}
\end{figure}

In Figures~\ref{rad-table-example} and \ref{slake-table-example}, we provide examples from the VQA-RAD and SLAKE datasets, respectively. Our goal is to illustrate the performance of the original MEVF and QCR in comparison with QCR when PubMedCLIP is used as the visual encoder for the MedVQA task. We refer to the QCR model with PubMedCLIP simply as PubMedCLIP in these figures. Examples from both datasets in Figures~\ref{rad-table-example} and \ref{slake-table-example} demonstrate that the MEVF model has difficulties correctly comprehending which organ is depicted in the image. For example, regardless of the asked question, in the left image in Figure~\ref{rad-table-example}, we observe that although the given image is a chest x-ray, the answer that MEVF provides is related to the abdominal region. This behaviour is seen for the right examples as well, where the given image is from the brain, but the predicted answer from MEVF relates to the chest. The same behaviour is further also observed in Figure~\ref{slake-table-example}. From this perspective, QCR appears to be providing answers that are at least relevant to the given image and question, although it fails to select the correct answer. As an example, for the left image in Figure~\ref{rad-table-example}, QCR understands the question as well as the image, but cannot provide the coarse correct answer. In the left image from Figure~\ref{slake-table-example},  QCR appears to understand that it relates to the lung/chest area, but it has apparent difficulties in comprehending the question and providing the correct answer. In contrast, using QCR with PubMedCLIP shows an improvement and results in providing answers that are correct throughout all examples in Figures~\ref{rad-table-example} and \ref{slake-table-example}.

\begin{figure}
\centering
  \includegraphics[width=\linewidth]{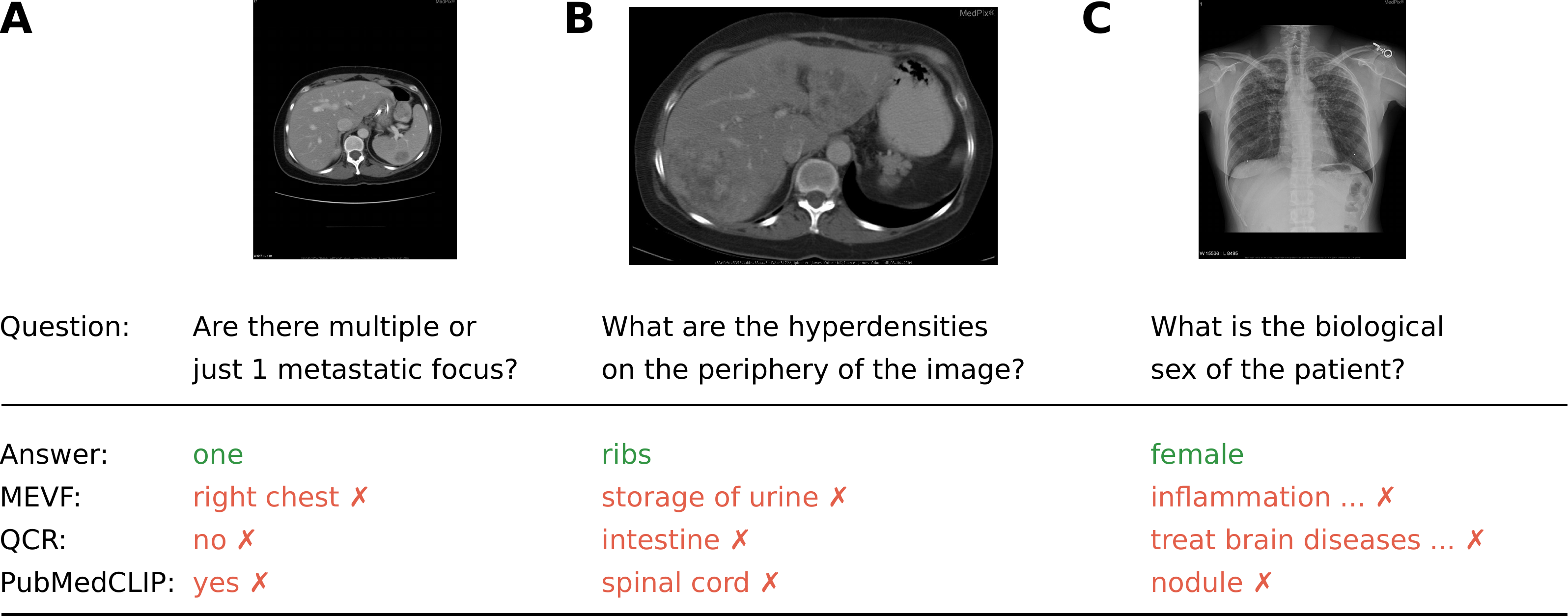}
 \caption{Examples from VQA-RAD where all models fail} \label{rad-resutls-fail}
\end{figure}

For further analysis, we provide examples in Figure~\ref{rad-resutls-fail} from the VQA-RAD dataset, where all three models fail to yield the correct answer. We again observe that MEVF provides irrelevant answers about body organs that are not present in the image. QCR shows the same behaviour for the right-most example. For the image on the left, QCR and PubMedCLIP miscomprehend the question as a yes/no question. In spite of this, the fact that PubMedCLIP answers with ``yes" illustrates that it has at least detected the ``one" metastatic focus in the image. In comparison, QCR answers with ``no", showing its troubles in interpreting the image and recognizing the metastatic focus. In the center example, answers provided by QCR and PubMedCLIP both appear to be relevant to the content of the given image. This suggests that the models have trouble understanding the semantics of the expression ``periphery of the image" in the question. By invoking techniques such as Grad-CAM \cite{gradCam}, we may be able to better understand what part of the image the model was focusing on before the classification layer. Finally, in the right-most example, QCR appears to misinterpret the content of the chest x-ray image and give suggestions for treating the brain. However, PubMedCLIP's answer, lung ``nodule", seems to be at least relevant to the image, although it shows that it is having difficulties inferring the semantics of the question. Our observations reveal that these models still have shortcomings in understanding questions and correctly relating them to the images.

\subsection{CLIP in MedVQA versus general-domain VQA}
Using CLIP in general-domain VQA, as investigated in prior work \cite{CLIPbenefit}, evinces its effectiveness in comparison to previous ResNet-based encoders. In such settings, it has been observed that CLIP with a ResNet visual encoder outperforms using Vision Transformers. The authors hypothesize that this is due to Vision Transformers' weakness in visual localization. Furthermore, their reports show that larger back-end visual encoders in CLIP such as ResNet-101 and ResNet-50x4 result in bigger gains in accuracy. 

In the MedVQA domain, we as well observe that PubMedCLIP outperforms the previous MAML-based visual encoders. However, using the bigger ResNet-50x4 model as the visual encoder in PubMedCLIP appears to lead to overfitting on medical images and it therefore performs slightly worse than the smaller version ResNet-50. Moreover, our experiments on the SLAKE dataset show that the Vision Transformer encoder in PubMedCLIP slightly outperforms using a ResNet-50. We showed that this gap stems mainly from differences in the underlying VQA data distributions. In the SLAKE dataset, the majority of the questions target a holistic overview of the image. In contrast, questions in VQA-RAD are primarily about the presence of an abnormality and require visual localization.
\section{CONCLUSION}
This work introduces PubMedCLIP as a pre-trained visual encoder for medical images and illustrates its effectiveness for the task of medical visual question answering. PubMedCLIP is trained using the image--caption pairs from thousands of PubMed articles. Our experiments on two benchmark MedVQA datasets demonstrate that PubMedCLIP outperforms previously used pre-trained visual encoders in MedVQA by up to $3\%$. Furthermore, our results reveal differences of underlying data distributions in the two benchmark datasets. We hope that our findings encourage future research to make real-world clinical image--text pairs publicly available for better development of vision--language representation learning with cross-modal supervision in medical domain. In terms of future work, further analysis of these models using explainable AI techniques such as Grad-CAM visualizations can enable us to assess their regions of focus within the image from the class activation maps. Moreover, by releasing PubMedCLIP, we hope to enable research investigating to what extent it may contribute to additional medical use-cases such as image classification for medical diagnosis and radiology report generation.

\section{ACKNOWLEDGMENT}
We would like to thank Matthias Steinbrecher for his helpful comments and discussions.

\section{CONFLICT OF INTEREST STATEMENT}
None.

\bibliographystyle{unsrtnat}
\bibliography{references}

\end{document}